\documentclass[5p,authoryear]{elsarticle}
\makeatletter 
\def\ps@pprintTitle{%
 \let\@oddhead\@empty
 \let\@evenhead\@empty
 \let\@evenfoot\@oddfoot} 
\makeatother

\usepackage[utf8]{inputenc} 
\usepackage[T1]{fontenc}
\usepackage[babel=true]{csquotes} 
\usepackage[fleqn]{amsmath} 
\usepackage{amsthm} 
\usepackage{booktabs} 
\usepackage{multirow} 
\usepackage{amssymb} 
\usepackage{float}
\usepackage{graphicx}
\usepackage{subfigure}
\usepackage{diagbox}

\newlength{\Oldarrayrulewidth}
\newcommand{\Cline}[2]{%
\noalign{\global\setlength{\Oldarrayrulewidth}{\arrayrulewidth}}%
\noalign{\global\setlength{\arrayrulewidth}{#1}}\cline{#2}%
\noalign{\global\setlength{\arrayrulewidth}{\Oldarrayrulewidth}}}

\usepackage{apalike}
\begin{document}

\begin{frontmatter}

\title{Improving Interpretability of Word Embeddings by Generating Definition and Usage}


\author[]{Haitong Zhang}
\ead{z962630523@gmail.com}

\author[]{Yongping Du\corref{cor1}}
\ead{ypdu@bjut.edu.cn}

\author[]{Jiaxin Sun}
\ead{sjx725224@163.com}

\author[]{Qingxiao Li}
\ead{lqx\_bjut@163.com}

\cortext[cor1]{Corresponding author.}
\address{Faculty of Information Technology, Beijing University of Technology, Beijing, 100124, China}

\begin{abstract}
	Word embeddings are substantially successful in capturing semantic relations among words. However, these lexical semantics are difficult to be interpreted. Definition modeling provides a more intuitive way to evaluate embeddings by utilizing them to generate natural language definitions of corresponding words. This task is of great significance for practical application and in-depth understanding of word representations. We propose a novel framework for definition modeling, which can generate reasonable and understandable context-dependent definitions. Moreover, we introduce \emph{usage modeling} and study whether it is possible to utilize embeddings to generate example sentences of words. These ways are a more direct and explicit expression of embedding's semantics for better interpretability. We extend the single task model to multi-task setting and investigate several joint multi-task models to combine usage modeling and definition modeling together. Experimental results on existing Oxford dataset and a new collected Oxford-2019 dataset show that our single-task model achieves the state-of-the-art result in definition modeling and the multi-task learning methods are helpful for two tasks to improve the performance.
	
\end{abstract}


\end{frontmatter}

\section{Introduction}\label{introduction}
Word embeddings \citep{turian2010word} have been exploited to obtain superior performance on many NLP tasks \citep{huang2014learning,tai2015improved,yang2016hierarchical}. Pre-trained embeddings \citep{mikolov2013efficient,pennington2014glove} learned from large-scale textual data have been shown to capture lexical syntax and semantics via word similarity \citep{landauer1997solution,downey2007sparse} or word analogy \citep{mikolov2013linguistic} tasks. However, these tasks can only evaluate the lexical information indirectly and lack sufficient interpretability. 

Considering that the dictionary definitions can represent the sense of words explicitly, \citet{noraset2017definition} introduces definition modeling to interpret an embedding's semantics by directly generating the textual definition of the corresponding word. The primary motivation is that if the vectors can capture abundant semantic information, readable definition of the word can be constructed from it. Definition modeling evaluates the semantics embedded in vectors more transparently than previous works with better interpretability. Especially, definition generation has a high practical value for various applications. It is helpful to provide definitions when people are reading or learning unknown words \citep{ishiwatari2019learning}. Traditional manual lexicography is time-consuming and laborious, while definition model can assist in dictionary compilation, especially for low-resource languages.

The distributional hypothesis \citep{harris1954distributional} indicates that words which occur in similar contexts tend to have similar meanings. In other words, the context of a word has the ability to represent the word sense. The context is also critical to solve the problem of word ambiguity in definition modeling. We argue that the semantics captured by word embeddings can reconstruct its context in reverse. Thus, we introduce the \emph{usage modeling} task to explore the possibility of using word embeddings to generate usages (example sentences). Given a target word and its embedding, \emph{usage modeling} estimates the probability of the usage for target word.

In this work, we focus on improving the interpretability of word embeddings by semantics generation. We first explore definition modeling and design a novel architecture based on the encoder-decoder framework. Unlike \citet{gadetsky2018conditional} which only uses the context for word sense disambiguation, the context is used to encode context-aware semantic representation of the target word by scaled dot-product attention \citep{vaswani2017attention} and directly guide token generation. We initialize the hidden state with the context embedding and provide the ELMo embeddings \citep{peters2018deep} of the target word for decoder to utilize contextual information explicitly during decoding. Inspired by multi-task sequence to sequence learning methods \citep{luong2015multi}, two joint multi-task models are proposed to combine usage modeling and definition modeling, sharing the representations at different levels. Compared with previous works, we employ multi-task settings to generate definition and example sentence for the target word simultaneously. These methods can be applied to create intelligent dictionary systems to help learning and reading in the intelligent education field. And the definitions and usages can be used as lexical knowledge to improve the performance of other tasks such as word sense disambiguation \citep{camacho2018word}.

Our contributions are as follows: (1) A new model is proposed to generate context-dependent definitions, which achieves the state-of-the-art result compared with the previous definition models regardless of considering context or not. Further, the character-level and contextualized word embeddings are used to compensate for the drawbacks of traditional word embeddings and provide better representations. (2) Usage modeling task is introduced, and two kinds of multi-task models are investigated that combine usage modeling and definition modeling by parallel-shared model and hierarchical-shared model respectively. To the best of our knowledge, our work is the first study trying to generate example sentences by neural networks. (3) We construct a large dictionary dataset, which provides a wealth of dictionary resources that can be leveraged in different NLP tasks.

\section{Related Work} 

Recently, many works utilize dictionary definitions to learn or improve word embeddings. \citet{wang2015learning} incorporate dictionary as a lexicographic knowledge to learn embeddings. \citet{hill2016learning} explore the representations of phrases and sequences by training neural networks to produce a target embedding for a word given its definition. \citet{bosc2018auto} improve word embeddings with an auto-encoder structure that begins from the target word embedding and backs to the definition. \citet{scheepers2018improving} utilize the definitions to tune word embeddings towards better compositionality.

Interpreting word embeddings is often conducted throu\-gh non-negative and sparse coding \citep{faruqui2015sparse,luo2015online,kober2016improving}, or regularization \citep{sun2016sparse}. In addition, \citet{park2017rotated} apply rotation algorithms based on exploratory factor analysis (EFA) to embedding spaces for obtaining interpretable dimensions in an unsupervised manner. These approaches rely on solving complex optimization problems, while we make semantic information of embeddings explicit by generating text.  

The most related work to this paper is definition modeling \citep{noraset2017definition} which is a task that estimates the probability of dictionary definition based on the defined word and its embedding. Several RNN-based definition models have been proposed, which are trained on the dictionary definitions corpus and then used to generate the corresponding definitions of the words not seen during training. However, they treat all of the words as monosemic words, which brings the problem of word ambiguity. To address the polysemy issue, \citet{gadetsky2018conditional} extend the context information of the defined word into definition modeling. They propose two definition models that incorporate context information. One of them is based on Adaptive Skip-gram vector representations \citep{bartunov2016breaking} that provides different word embeddings according to its contexts. The other uses soft attention mechanism to extract the components of word embedding relevant to corresponding meaning. \cite{chang2018xsense} incorporate a sparse vector extractor to select sense-specific representation of the target word, which aims to provide a more explainable model for definition generation. In addition, \citet{yang2019incorporating} incorporate sememes into Chinese definition modeling.

\section{Usage Modeling}

Word embeddings is usually learned from words co-occurrence information in large corpus based on the distributional hypothesis \citep{mikolov2013efficient,pennington2014glove}. The distribution hypothesis \citep{harris1954distributional} holds that words occurring in similar contexts are semantically similar. In turn, the semantic information captured by the embeddings should be able to reconstruct the contexts. We introduce usage modeling to further improve the interpretability of word embeddings by generating the usage of the target word. Usage generation is similar to definition generation but with some differences. For definition generation, words other than the defined word are used to express the meaning of the target word. While, for usage generation, the model must understand how to use the target word in the context. This is a more profound and more complex lexical semantic representation of words.

 Usage modeling estimates the probability of words in usage sentence $U=\left\{w_{1}, \dots, w_{N}\right\}$ given a target word $w^{*}$. The corresponding context $C=\left\{c_{1}, \dots, c_{m}\right\}$ is also considered to avoid the problem of ambiguity. The joint probability is decomposed into separate conditional probabilities as:
\begin{equation}
p\left(U | w^{*}, C\right)=\prod_{n=1}^{N} p\left(w_{n} | w_{i<n}, w^{*}, C\right)
\end{equation}
where $i=\left\{1, \dots, N\right\}$ denotes the index of word in usage sentence. Note that the target word should be included in the usage sentence. Similarly, usage modeling is a special case of language modeling, and the performance can be measured by the perplexity of test data. Below is an example entry from the Oxford English Dictionary:

\textbf{Soldier:} \emph{A person who serves in an army.}

\textbf{example sentence:} \emph{He zoomed in on the view of one of the soldiers under his command.}

Dictionary definitions are usually comprised of {\em genus\/} and {\em differentiae\/} \citep{chodorow1985extracting, montemagni1992structural}, where the {\em genus\/} is the hypernym of the defined word and the {\em differentiae\/} distinguishes the hypernym from the defined word. The word ``soldier'' in the above example is defined as ``a person who serves in an army'', where the ``person'' is genus and the rest is differentiae. However, the example sentences do not have a specific structure like definitions and there are a wide variety of tenses and lexical forms in sentences such as the word in the example above, from singular ``soldier'' to plural ``soldiers''. These make usage modeling more complicated, which requires a deeper understanding of semantics and syntax.

\begin{figure*}[t]
	\centering  
	\includegraphics[scale=0.57]{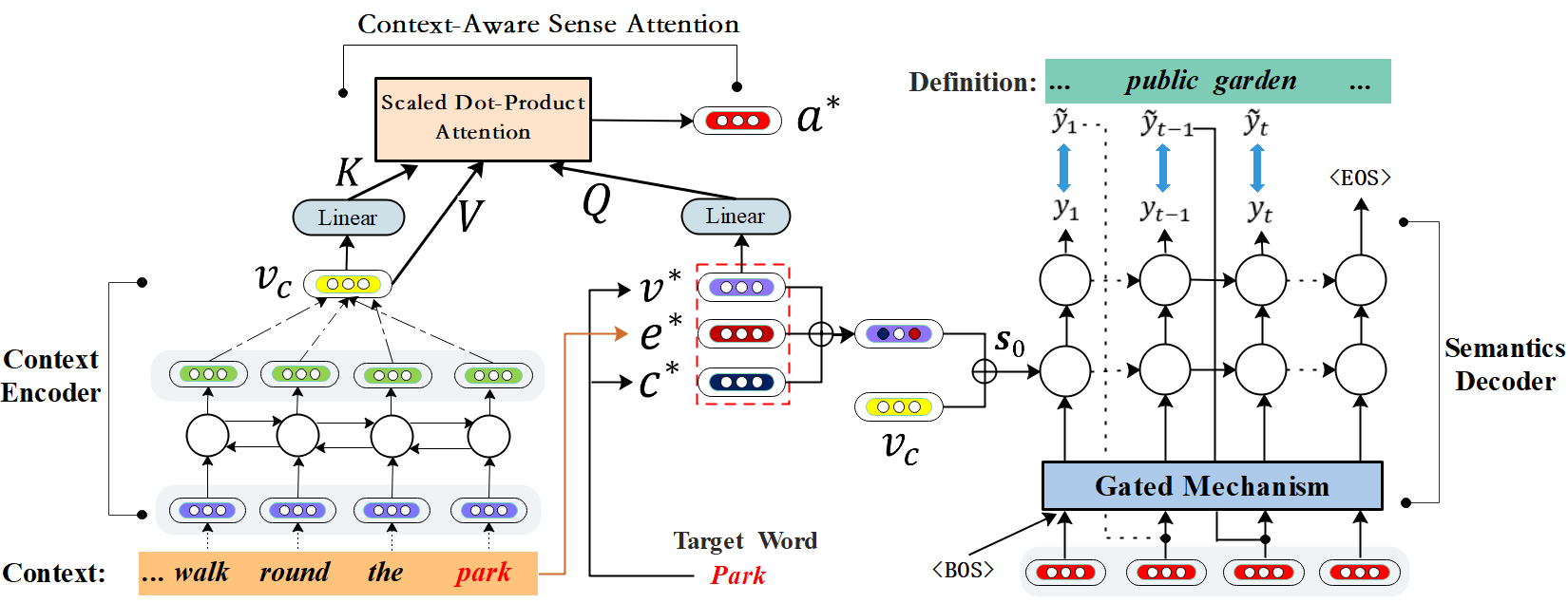}  
	\caption{An overview of the definition model Semantics-Generator. The context encoder maps the context into a sentence embedding and the scaled dot-product attention is used to generate the context-aware word embedding. Character-level embeddings and ELMo are used as additional features and the decoder generates context-dependent definition for the target word.
	}  
	\label{fig1}   
\end{figure*}

\section{A New Oxford-2019 Dataset}
Rich dictionary resources are accessible online, such as WordNet, Wiktionary and Merriam-Webster Dictionary. However, most of them are short of example sentences compared with Oxford Dictionary. \citet{gadetsky2018conditional}
collected a dataset utilizing Oxford Dictionary resources. As described in \cite{chang2018xsense}, there is some noise in the data, and the format is messy where the defined word contains {\em Arabic numerals\/}, {\em Non-English alphabets\/} and {\em Special symbols\/}. Moreover, some example sentences do not contain the target word which makes the context useless. To make up for these deficiencies and explore usage modeling, we collect a new Oxford-2019 dataset by Oxford Dictionaries API\footnote{https://developer.oxforddictionaries.com/}.

We sample the most common 65,000 tokens in WikiText-103 dataset \citep{merity2016pointer}, removing duplicate words, function words and stop words, keeping only the words that consist of pure letters as the vocabulary. The ``sentences'' returned by Oxford Dictionaries API are used as contexts while the ``examples'' are used as usages so that we can model example sentence and distinguish corresponding meaning by the context. Here, ``sentences'' are extracted from corpora, using richer vocabulary and longer sentences than ``examples''. Each specific definition includes three different context sentences and one usage sentence. The collected data are split into train, valid, and test sets by specific meaning of words so that the same definition for a given word does not appear in different sets.

The statistics of Oxford-2019 dataset are shown in Table \ref{table1}, which contains example sentences with more contexts, and the size is much larger than that provided by \citet{gadetsky2018conditional}. The domain information that may provide knowledge of the proper genus and POS tag of the word are also included, which could be utilized by other NLP tasks for further research.

\begin{table}[]
	\centering
	\begin{tabular}{c||r|r|r}
		\hline
		\textbf{Split} & \textbf{Train} & \textbf{Valid} & \textbf{Test} \\ \hline
		\#Words & 32,066 & 7,359 & 7,322 \\
		\#Entries & 234,949 & 29,275 & 29,241 \\
		\#Tokens & 2,595,561 & 323,700 & 322,915 \\ \hline
		Def Len & 11.05 & 11.06 & 11.04 \\
		Ctx Len & 21.48 & 21.43 & 21.43 \\
		Usg Len & 11.30 & 11.28 & 11.31 \\ \hline
	\end{tabular}
	\caption{The basic statistics of the new Oxford-2019 dataset. The average length of the contexts is much longer ($>$10) than the usages.}
	\label{table1}
\end{table}

\section{Models}
We introduce the proposed definition model \textbf{Semantics-Generator} as shown in Figure \ref{fig1}. Further, two kinds of multi-task models are proposed combining usage modeling and definition modeling.

\subsection{Semantics-Generator: Semantic Generative Network for Word Embedding }

\textbf{Context Encoder.} The context encoder aims at encoding the context sequence into continuous vectors and generating a meaningful sentence embedding of fixed-size. We adopt bidirectional GRU with a max pooling layer as the encoder architecture\citep{conneau2017supervised}. Given a context containing $m$ words $C=\left\{c_{1}, \dots, c_{m}\right\}$, bidirectional GRU computes a set of $m$ vectors $\boldsymbol{C}_{v} = \left\{\boldsymbol{h}_{i}\right\}_{m}$ that each $\boldsymbol{h}_{i} \in\left(\boldsymbol{h}_{1}, \ldots, \boldsymbol{h}_{m}\right)$ is the concatenation of the feed-forward direction and the back-forward direction respectively:

\begin{equation}\label{eq:12}
\begin{split}
& \overrightarrow{\boldsymbol{h}_{i}}=\overrightarrow{\operatorname{G R U}_{i}}\left(\boldsymbol{c}_{1}, \ldots, \boldsymbol{c}_{m}\right) \\
& \overleftarrow{\boldsymbol{h}_{i}}=\overleftarrow{\operatorname{G R U}_{i}}\left(\boldsymbol{c}_{1}, \ldots, \boldsymbol{c}_{m}\right) \\
& \boldsymbol{h}_{i}=\left[\overrightarrow{\boldsymbol{h}_{i}}, \overleftarrow{\boldsymbol{h}_{i}}\right]
\end{split}
\end{equation}
where $\boldsymbol{h}_{i} \in \mathbb{R}^{2d_{h}}$, $d_{h}$ is the hidden size of the encoder and $\boldsymbol{c}_{i}\in \mathbb{R}^{d_{w}}$ denotes the word embedding of $c_{i}$. Then, a max pooling layer is used to select the maximum value over the hidden units for each dimension to get a vector $\boldsymbol{v}_{c}\in \mathbb{R}^{2d_{h}}$ as the embedded representation of the context.

\textbf{Context-Aware Sense Attention.} \citet{gadetsky2018conditional} use a soft binary mask which depends on the context to filter out unrelated components in the target word embedding for context-aware definition generation. However, the context length tends to be long in the datasets, as they just treat all words in the context indiscriminately which may bring some noises in disambiguating relevant meaning. Considering the sense of the target word depends on some relevant words in context and the context information may directly benefits the generation process, we use attention mechanism to generate a new target word representation aligned to its context. 

Self-attention has the ability to well capture the long distant dependency information and it does not create much computational complexity, which improves the performance on many NLP tasks \citep{vaswani2017attention}. We modify its scaled dot-product attention to compute the context-aware word embedding. First, the target word embedding $\boldsymbol{v}^{*}$ is mapped to a matrix $\boldsymbol{Q} \in \mathbb{R}^{1 \times d}$ and its context is mapped to key-value pairs $\boldsymbol{K} \in \mathbb{R}^{m \times d}$ and $\boldsymbol{V} \in \mathbb{R}^{m \times d}$ by using different linear projections. Then, we compute the dot products of the target word with all keys of the context, divided each by $\sqrt{d}$, and apply a softmax function to obtain the weights on the values. Finally, a linear operation is applied to get the output. The equations can be represented as follows:

\begin{gather}
\boldsymbol{Q}=\boldsymbol{v}^{*} \boldsymbol{W}^{Q},
\boldsymbol{K}=\boldsymbol{C}_{v} \boldsymbol{W}^{K},
\boldsymbol{V}=\boldsymbol{C}_{v} \boldsymbol{W}^{V} \\
\operatorname{Attention}\left(\boldsymbol{Q}, \boldsymbol{K},\boldsymbol{V}\right)=\operatorname{softmax}\left(\frac{\boldsymbol{Q} \cdot \boldsymbol{K}^{\top}}{\sqrt{d}}\right) \boldsymbol{V} \\
\boldsymbol{a}^{*}=\boldsymbol{W}^{O} \cdot \operatorname{Attention}\left(\boldsymbol{Q}, \boldsymbol{K},\boldsymbol{V}\right)\end{gather}
where $\boldsymbol{W}^{Q} \in \mathbb{R}^{d \times d_{w}}$, $\boldsymbol{W}^{K} \in \mathbb{R}^{d \times 2d_{h}}$, $\boldsymbol{W}^{V} \in \mathbb{R}^{d \times 2d_{h}}$, $\boldsymbol{W}^{O} \in \mathbb{R}^{d_{w} \times d}$ are all trainable parameter matrices, and $d$ is the number of the hidden units in the model. The attention mechanism dynamically incorporates matching information from the context to generate a context-aware word representation $\boldsymbol{a}^{*}$, which is used to provide relevant sense to the decoder at each time step.

\begin{figure*}[t]
	\centering

	\quad
	\subfigure[Parallel-Shared Model.]{
		\includegraphics[width=7cm]{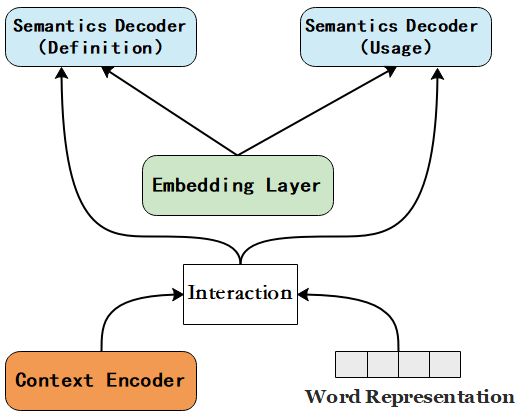}\label{fig2a}
	}
	\quad
	\subfigure[Hierarchical-Shared Model.]{
		\includegraphics[width=7cm]{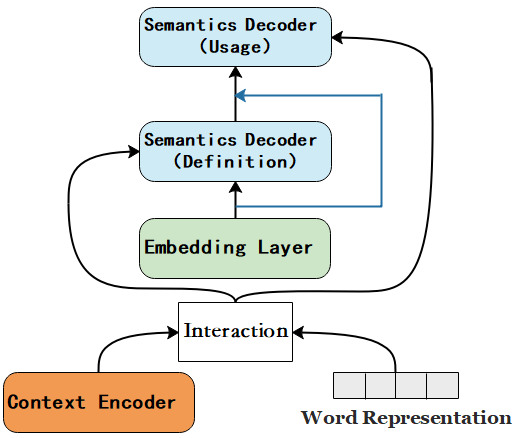}\label{fig2b}
	}
	\caption{Multi-task models by parallel and hierarchical structure. Parallel-Shared Model shares the representations at the same level while Hierarchical-Shared Model shares the representations at hierarchical level.}\label{fig2}
\end{figure*}
\textbf{Semantics Decoder.} Given the target word $w^{*}$ with its context $C$, another GRU decoder \citep{cho2014learning} defines the probability over the definition $D=\left\{y_{1}, \dots, y_{T}\right\}$ as a joint probability of ordered conditionals:
\begin{gather}
p\left(D | w^{*}, C\right)=\prod_{t=1}^{T} p\left(y_{t} | y_{i<t}, w^{*}, C\right) \\
p\left(y_{t} | y_{i<t},w^{*}, C \right)=\operatorname{softmax}\left(\boldsymbol{W}_{d} \boldsymbol{s}_{t}+\boldsymbol{b}_{d}\right)\end{gather}
where $\boldsymbol{W}_{d} \in \mathbb{R}^{\left|Y\right| \times d_{s}}$, $\boldsymbol{b}_{d} \in \mathbb{R}^{\left|Y\right|}$ is a linear projection to a vocabulary-sized vector and bias term respectively, $\boldsymbol{s}_{t}$ is the hidden state of the decoder GRU at time step $t$ and $d_{s}$ is the hidden size of decoder. Particularly, we assign the concatenation of the target word embedding $\boldsymbol{v}^{*}$ and its context embedding $\boldsymbol{v}_{c}$ to the initial hidden state $\boldsymbol{s}_{0}$, which can provide more comprehensive and explicit signal of the target word and the context for the decoder to generate smooth and consistent context-dependent definitions:
\begin{equation}
\boldsymbol{s}_{0}=\boldsymbol{W}_{s}\left([\boldsymbol{v}^{*} ; \boldsymbol{v}_{c}]+\boldsymbol{b}_{s}\right) \label{eq:initial_state}
\end{equation}
 Here, $[a ; b]$ denotes vector concatenation, $\boldsymbol{W}_{s} \in \mathbb{R}^{d_{s} \times \left(d_{w}+2d_{h}\right)}$ and $\boldsymbol{b}_{s} \in \mathbb{R}^{d_{s}}$ are parameter matrix and bias term respectively. 

Each word in the definition has a varying degree of dependency on the defined word. Moreover, the $t$-th generated word $y_{t}$ also depends on previous words $y_{1}, \dots, y_{t-1}$ variously. Note that the decoder is autoregressive during the generation process. In order to focus on some features which are more useful to generate each word and alleviate the influence of previous generated error words on the subsequent decoding process, we add a gated function to the input $\boldsymbol{x}_{t}$ at each decoding step as:
\begin{align}
\boldsymbol{x}_{t}&=\boldsymbol{g}_{t} \odot [\boldsymbol{a}^{*} ; \boldsymbol{y}_{t-1};\boldsymbol{c}^{*};\boldsymbol{e}^{*}] \\
\boldsymbol{g}_{t}&=\sigma(\boldsymbol{W}_{g}  [\boldsymbol{a}^{*} ; \boldsymbol{y}_{t-1};\boldsymbol{c}^{*};\boldsymbol{e}^{*} ] ) \\
\boldsymbol{c}^{*}&=\operatorname{CNN}(w^{*}) \\
\boldsymbol{e}^{*}&=\operatorname{ELMo}(C)
\end{align}
where $\boldsymbol{W}_{g}$ is the projection matrix, $\sigma(\cdot)$ and $\odot$ denote the sigmoid function and element-wise vector multiplication operation respectively. $\boldsymbol{c}^{*}$ and $\boldsymbol{e}^{*}$ are namely the character-level information and the ELMo embeddings \citep{peters2018deep} of the target word, which are explained in detail later. Unlike \citet{noraset2017definition}, our gated function combines the features of the target word dynamically, and controls the influence of both the defined word and the current input at each time step. Then, the GRU updates the output of each time step by:
\begin{equation}
\boldsymbol{s}_{t}=\operatorname{GRU}\left(\boldsymbol{s}_{t-1},\boldsymbol{x}_{t} \right)
\end{equation}
By providing explicit context signal and ELMo representations, the decoder can pay attention to the entire context information when decoding each token.

\textbf{Character-Level Information.}
Many words in English consist of a root and affixes. For example, the English morpheme \textbf{re-} often refers to again as ``\emph{reinforce}'' while \textbf{dis-} usually means opposite as ``\emph{dislike}''. These character-level morphological and semantic information have a certain impact on the meaning of the defined word, so the character-level embeddings are used to capture these features. Following \cite{noraset2017definition}, we utilize a character-level convolution neural network (CNN) followed by a max pooling layer to create these features \citep{kim2016character} as $\boldsymbol{c}^{*}$. Moreover, two-layer Highway Network \citep{srivastava2015highway} is used to solve training difficulties and generate better representation.

\textbf{Contextual Word Embeddings.}
Traditional word vectors only allow a single context-independent representation for each word which is difficult to represent the complex characteristics of a word and its varied semantics in different contexts. We use pre-trained ELMo embeddings \citep{peters2018deep} to supplement the contextualized word representation, denoted as $\boldsymbol{e}^{*}$. ELMo is trained from the hidden states of a deep bidirectional language model so that one word may have different representations depending on the corresponding contexts.

\subsection{Multi-Task Learning with Usage Modeling}
Multi-task sequence to sequence learning \citep{luong2015multi} has achieved excellent results on many sequence modeling tasks \citep{niehues2017exploiting,clark2018semi}. In order to further improve the interpretability of word embeddings, two kinds of multi-task models that combine usage modeling and definition modeling are proposed, sharing the representations at different levels by parallel and hierarchical level respectively, as shown in Figure \ref{fig2}. For a given word and its context, our multi-task models can generate both definition and example sentence of the target word. Since definition modeling and usage modeling are based on the same target word and context, the multi-task models have similar architecture as Semantics-Generator, with just one more decoder for usage modeling.

\textbf{Parallel-Shared Model.} In parallel-shared model, both definition modeling and usage modeling are supervised at the same-level layer. For each task, we use two separate task-specific semantics decoders only sharing the embedding layer. Both decoders receive the concatenation of the target word embedding $\boldsymbol{v}^{*}$ and the context embedding $\boldsymbol{v}_{c}$ as the initial hidden state (Eq. \ref{eq:initial_state}), and compute their outputs respectively as follows:

\begin{equation}
\boldsymbol{s}_{t}^{d}=\operatorname{GRU}^{\emph{Def}}\left(\boldsymbol{s}_{t-1}^{d},\boldsymbol{x}_{t}^{y}\right)
\end{equation}
\begin{equation}
\boldsymbol{s}_{t}^{u}=\operatorname{GRU}^{\emph {Usg}}\left(\boldsymbol{s}_{t-1}^{u},\boldsymbol{x}_{t}^{w}\right)\label{sameu}
\end{equation}
where $\boldsymbol{x}_{t}^{y}$ and $\boldsymbol{x}_{t}^{w}$ refer to the input representation at each time step based on the definition and usage, respectively.

\textbf{Hierarchical-Shared Model.} Previous work suggests that there is sometimes a natural order among different tasks \citep{hashimoto2016joint}. We also propose two hierarchical multi-task models where definition modeling and usage modeling are supervised at different-level layers. Hir-Shared-DU supervises definition modeling at the bottom layer while usage modeling at the top layer and Hir-Shared-UD is the opposite. The shortcut connection is used so that the higher-level decoder can have access to encoding richer representations. For Hir-Shared-DU (Figure \ref{fig2b}), the input of higher-level decoder is the concatenation of the lower layer decoder output and the words representations, and the output is changed from Eq. \ref{sameu} to the following:
\begin{equation}
\boldsymbol{s}_{t}^{\prime}=\operatorname{GRU}^{\emph {Def}}\left(\boldsymbol{s}_{t-1}^{\prime},\boldsymbol{x}_{t}^{w}\right)
\end{equation}
\begin{equation}
\boldsymbol{s}_{t}^{u}=\operatorname{GRU}^{\emph {Usg}}\left(\boldsymbol{s}_{t-1}^{u},\boldsymbol{W}_{p} \left[\boldsymbol{x}_{t}^{w} ; \boldsymbol{s}_{t}^{\prime}\right]\right)
\end{equation}
where $\boldsymbol{W}_{p}$ is a weight matrix to fuse the word representation and lower-level output into a common space.

\section{Experimental Setup}

\subsection{Datasets and Automatic Evaluation Metrics}
We train and evaluate our proposed definition model Semantics-Generator on both the original Oxford dataset constructed by \citet{gadetsky2018conditional} and our new collected Oxford-2019 dataset. As can be seen from Table \ref{table2}, the original Oxford dataset contains 122k sentences and 3.5M tokens. The size of the three sets are 97k, 12k, and 12k respectively. For the multi-task models, we report the performance on our new dataset.

We use BLEU \citep{papineni2002bleu} and ROUGE \citep{lin2004rouge} score as automatic evaluation metrics. Follow previous work \citep{noraset2017definition,gadetsky2018conditional}, the average BLEU score is computed across all test entries by the sentence-BLEU binary in Moses library\footnote{http://www.statmt.org/moses/} for fair comparison. Moreover, we report the average F-measure of ROUGE-L which is calculated by Rouge package\footnote{https://github.com/pltrdy/rouge}.
\begin{table}[]
	\centering
	\begin{tabular}{c||r|r|r}
		\hline
		\textbf{Split} & \textbf{Train} & \textbf{Valid} & \textbf{Test} \\ \hline
		\#Words & 33,128 & 8,867 & 8,850 \\
		\#Entries & 97,855 & 12,232 & 12,232 \\
		\#Tokens & 1,078,828 & 134,486 & 133,987 \\ \hline
		Def Len & 11.03 & 10.99 & 10.95 \\
		Ctx Len & 17.74 & 17.80 & 17.56 \\ \hline
	\end{tabular}
	\caption{The statistics of the Oxford dictionary dataset \citep{gadetsky2018conditional}.}
	\label{table2}
\end{table}
\subsection{Baselines}
We compare our models with two types of baselines that consider contexts or not.
The baselines that do not consider contexts \citep{noraset2017definition} are conditional RNN language models (RNNLM) based on the target word embedding.
The target word is added at the beginning of the definition as a form of \textbf{Seed} information \citep{sutskever2011generating}. The baselines with contexts include \textbf{S+I-Adaptive} and \textbf{S+I-Attention} which proposed by \citet{gadetsky2018conditional}. We reimplement these models with the same settings on the datasets for fair comparison.

\subsection{Experimental Details}

All models utilize the two-layer GRU network as the RNN component of the decoder, and word embeddings are initialized with pre-trained Word2Vec\footnote{https://code.google.com/archive/p/word2vec/} which is fixed during training. We search the hyper-parameters based on the perplexity scores on the valid set. Both word embeddings and GRU hidden layers of the decoder have 300 units. A one-layer bidirectional GRU with 150 hidden units is used for the context encoder. The context encoder has own embedding layer initialized by Word2Vec which is fine-tuned during training. For character-level embeddings, the CNN has size \{10, 30, 40, 40 ,40\} with kernels length 2 to 6. For ELMo embeddings, we use the pre-trained embeddings in size of 1024\footnote{https://allennlp.org/elmo}.

For our definition model Semantics-Generator, we first pre-train the semantics decoder by WikiText-103 dataset \citep{gadetsky2018conditional}. Both $\boldsymbol{v}^{*}$ and $\boldsymbol{s}_{0}$ are set to zero vectors for purely unconditional. Then, we train the full model in the definition dataset and minimize the negative log likelihood objective using Adam \citep{kingma2014adam}.
On the other hand, the multi-task models are trained directly in our definition corpus without pre-training, and the sum of negative log likelihood objective of two tasks is optimized. We only use the models with the lowest perplexity for generation by simple temperature sampling algorithm ($\tau=0.05$). Our source code and dataset are freely available\footnote{https://github.com/Haitons/Semantics-Generator}.
\begin{figure}[h]
	\centering
	\includegraphics[scale=0.5]{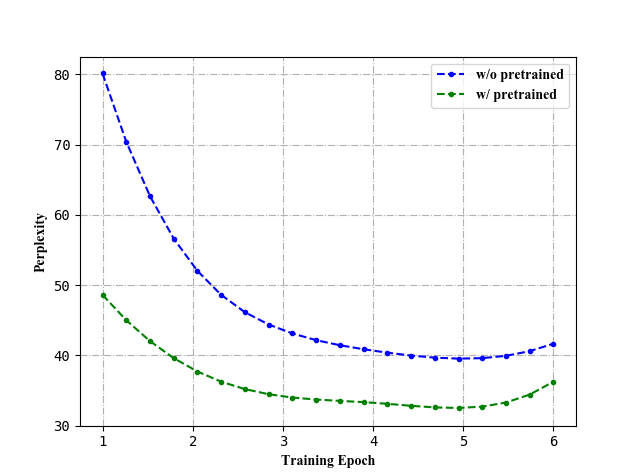}
	\caption{Perplexities of Semantics-Generator with or w/o pre-training on the Oxford dataset. Pre-training procedure can decrease perplexity effectively.}\label{fig3}
\end{figure}

\begin{table*}[]
	\centering
	\resizebox{\textwidth}{!}{%
		\begin{tabular}{lllclclcllclclc}
			\toprule[1.2pt]
			&                                                       & \multicolumn{6}{c}{\multirow{2}{*}{\textbf{Oxford Dataset }}}                                                     &                               & \multicolumn{6}{c}{\multirow{2}{*}{\textbf{Oxford-2019 (Our New Dataset)}}}                                          \\
			&                                                       & \multicolumn{6}{c}{}                                                                                            &                               & \multicolumn{6}{c}{}                                                                                            \\ \cline{1-8} \cline{10-15}
			\multirow{2}{*}{\textbf{Setup}} & \multicolumn{1}{l}{\multirow{2}{*}{\textbf{Methods}}} & \multicolumn{2}{c}{\textbf{Full}}   & \multicolumn{2}{c}{\textbf{Seen}}   & \multicolumn{2}{c}{\textbf{Unseen}} & \multicolumn{1}{c}{\textbf{}} & \multicolumn{2}{c}{\textbf{Full}}   & \multicolumn{2}{c}{\textbf{Seen}}   & \multicolumn{2}{c}{\textbf{Unseen}} \\
			& \multicolumn{1}{c}{}                                  & BLEU  & \multicolumn{1}{l}{ROUGE-L} & BLEU  & \multicolumn{1}{l}{ROUGE-L} & BLEU  & \multicolumn{1}{l}{ROUGE-L} &                               & BLEU  & \multicolumn{1}{l}{ROUGE-L} & BLEU  & \multicolumn{1}{l}{ROUGE-L} & BLEU  & \multicolumn{1}{l}{ROUGE-L} \\ \hline \hline
			1)                              & Models w/o contexts                            &       & \multicolumn{1}{l}{}        &       & \multicolumn{1}{l}{}        &       & \multicolumn{1}{l}{}        &                               &       & \multicolumn{1}{l}{}        &       & \multicolumn{1}{l}{}        &       & \multicolumn{1}{l}{}        \\
			A                               & \textit{Seed} \citep{noraset2017definition}                                        & 12.28 & 16.03                       & 12.54 & 15.71                       & 11.04 & 17.65                       & \multicolumn{1}{c}{}          & 12.55 & 18.01                       & 12.60 & 17.60                       & 12.37 & 19.28                       \\
			B                               & \textit{S+I}        \citep{noraset2017definition}                                   & 12.10 & 16.34                       & 12.38 & 16.11                       & 10.74 & 17.49                       & \multicolumn{1}{c}{}          & 12.80 & 18.53                       & 12.82 & 17.97                       & 12.75 & 20.14                       \\
			C                               & \textit{S+H}    \citep{noraset2017definition}                                       & 12.22 & 15.91                       & 12.45 & 15.58                       & 11.16 & 17.55                       & \multicolumn{1}{c}{}          & 12.64 & 18.42                       & 12.61 & 17.86                       & 12.72 & 20.13                       \\
			D                               & \textit{S+G}       \citep{noraset2017definition}                                    & 12.30 & 16.56                       & 12.47 & 16.24                       & 11.35 & 18.13                       & \multicolumn{1}{c}{}          & 12.77 & 18.53                       & 12.72 & 17.95                       & 12.89 & 20.31                       \\
			E                               & \textit{S+G+CH}    \citep{noraset2017definition}                                   & 12.36 & 16.77                       & 12.51 & 16.35                       & 11.59 & 18.88                       & \multicolumn{1}{c}{}          & 13.11 & 19.14                       & 13.02 & 18.48                       & 13.41 & 21.17                       \\ \hline 
			2)                             & Models w/ contexts                         &       & \multicolumn{1}{l}{}        &       & \multicolumn{1}{l}{}        &       &                             &                               &       &                             &       &                             &       &                             \\
			F                               & \textit{S+I-Adaptive}   \citep{gadetsky2018conditional}                              & 12.42 & 16.51                       & 12.63 & 16.13                       & 11.38 & 18.42                       &                               & 12.53 & 18.20                       & 12.47 & 17.54                       & 12.70 & 20.23                       \\
			G                               & \textit{S+I-Attention}   \citep{gadetsky2018conditional}                                & 12.36 & 16.54                       & 12.61 & 16.28                       & 11.08 & 17.84                       &                               & 12.31 & 18.37                       & 12.40 & 18.03                       & 12.05 & 19.41                       \\ \hline 
			H                               &  Ours: Semantics-Generator                                 & 15.73 & 21.56                       & 16.22 & 21.58                       & 13.30 & 21.43                       &                               & \textbf{13.85} & \textbf{20.67}                       & \textbf{13.91} & \textbf{20.30}                       & \textbf{13.68} & \textbf{21.83}                       \\I                               &  Ours: Semantics-Generator (Pre-train)                          & \textbf{16.00} & \textbf{21.94}                       & \textbf{16.53} & \textbf{21.98}                       & \textbf{13.38} & \textbf{21.70}                       &                               & 13.71 & 20.62                       & 13.77 & 20.29                       & 13.55 & 21.65                       \\ 
			\bottomrule[1.2pt]
		\end{tabular}%
	}
	\caption{The performance of Semantics-Generator and all baselines in definition generation on the test sets. We report the average scores across 5 trials.}
	\label{table3}
\end{table*}
\begin{figure*}[t]
	\begin{minipage}[t]{0.5\linewidth}
		\centering
		\includegraphics[width=3.2 in]{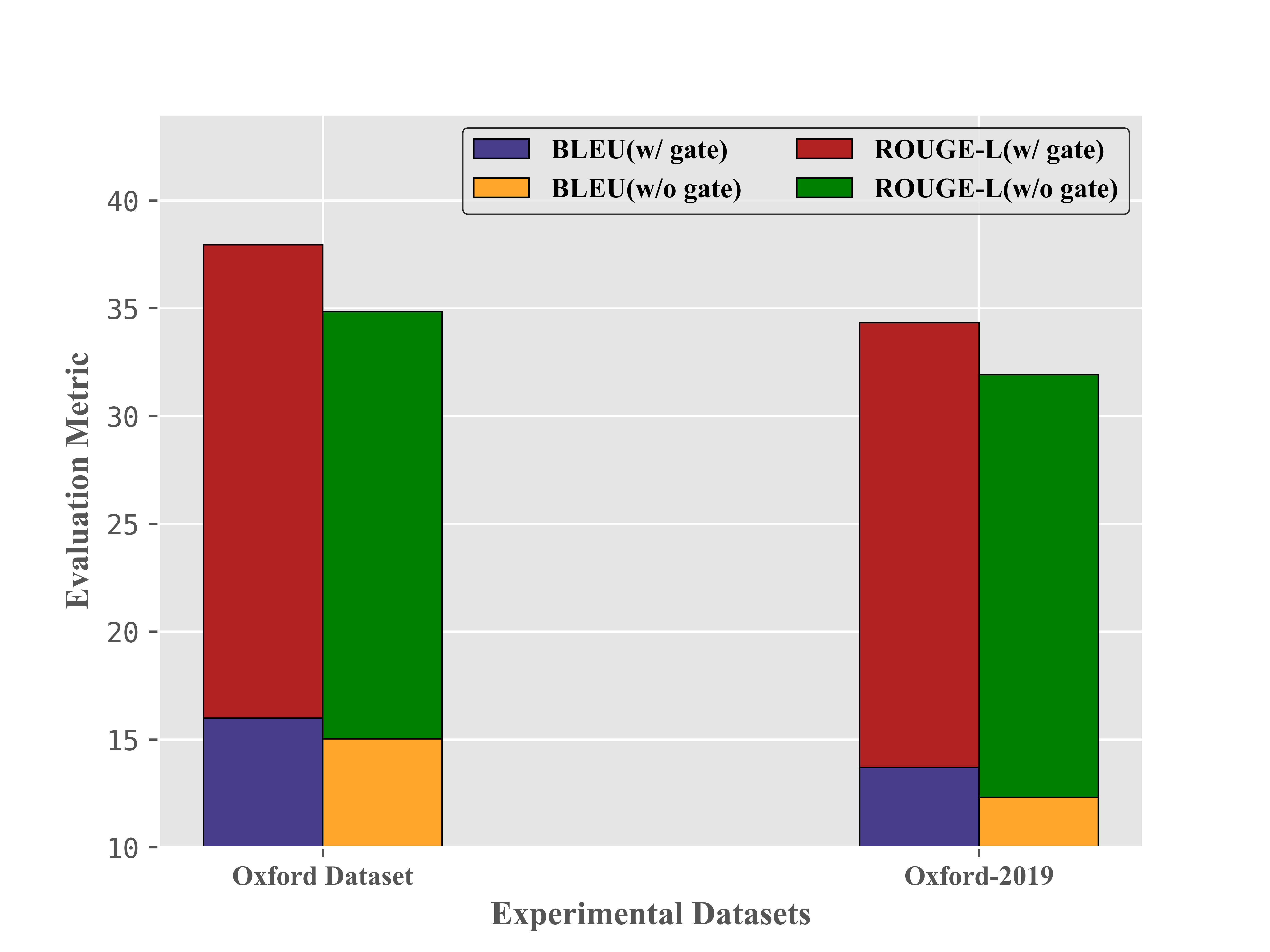}
		\caption{Evaluation metrics of with or w/o gated function. }
		\label{fig4}
	\end{minipage}%
	\begin{minipage}[t]{0.5\linewidth}
		\centering
		\includegraphics[width=3.2 in]{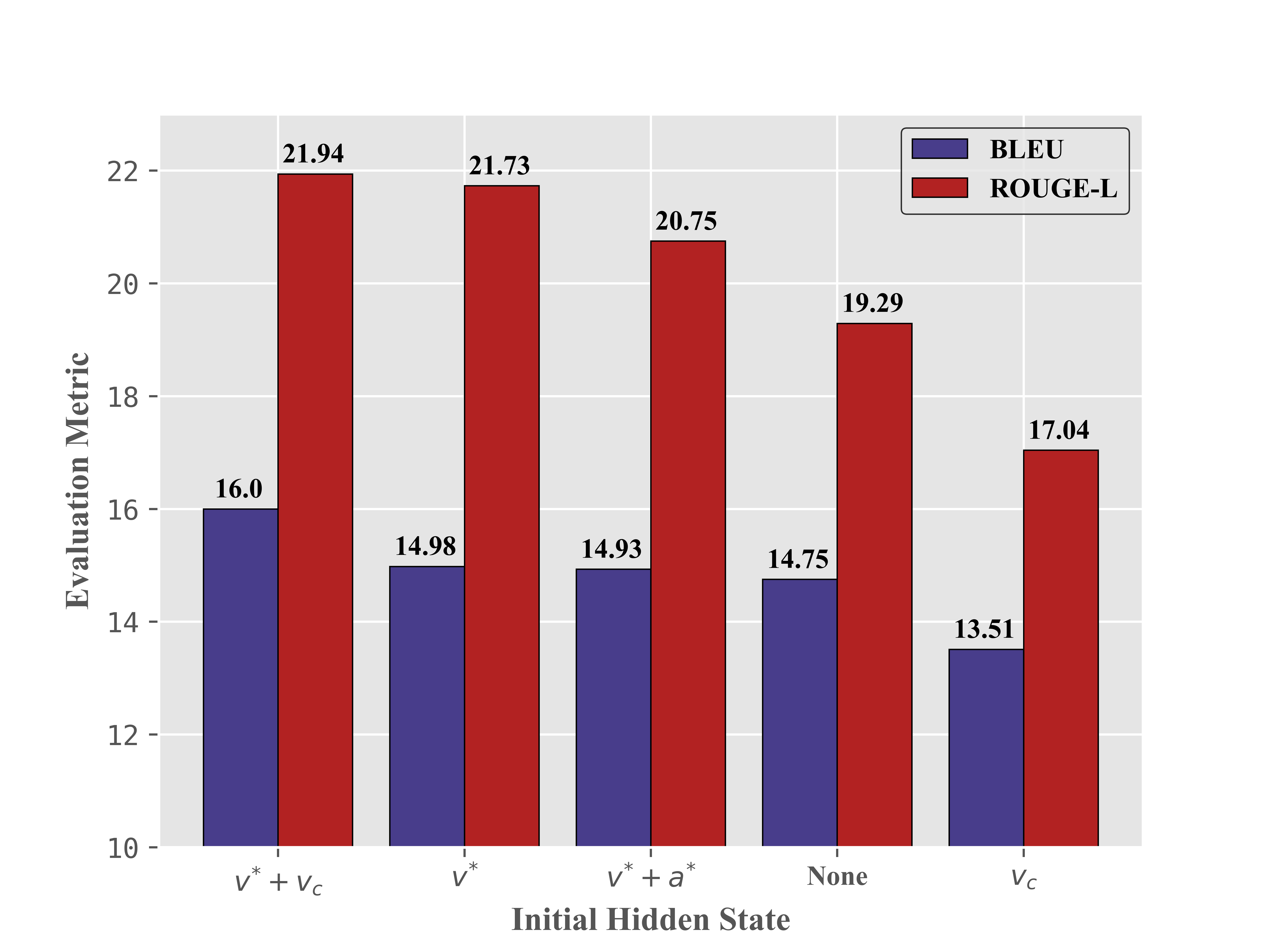}
		\caption{Evaluation metrics with varied initial hidden state on Oxford dataset.}
		\label{fig5}
	\end{minipage}
\end{figure*}
\section{Results and Discussion}
\subsection{Overall Performance}

\textbf{Automatic Evaluation} The performance of our definition model \textbf{Semantics-Generator} and all baselines (Model A-G) are reported in Table \ref{table3}. Compared with our new dataset, the original Oxford dataset is small which contains fewer entries. It is noticed that these datasets are split by specific definition, in which some test words may appear during training. We report the model performance on two different types of test data which are labeled as \textbf{Seen} and \textbf{Unseen} separately \citep{chang2018xsense}. The dataset with \textbf{Seen} label contains the target words with different meanings during training. On the contrary, the target words do not appear in the training set labeled as \textbf{Unseen}, which is more difficult to achieve the better performance. Among all of the baselines, the \textbf{S+I-Adaptive} model performs the best on the original small dataset (Model F). Although \citet{noraset2017definition}'s methods (Model A-E) do not take into account the contexts, their models still perform strong. On the small dataset, their best model \textbf{S+G+CH} performs slightly lower in BLEU, but achieves the better ROUGE-L (Model E vs. F\&G). Further, the performance of their work on our large dataset is improved and better than the baselines considering the contexts (Model A-E vs. F\&G). The probable reason is that since dictionary definitions usually have inherent structure, generating the same definition for all contexts can still have some common expressions and it may reduce the errors produced by the interaction between words and contexts. Our \textbf{Semantics-Generator} (Model H) gets a significant improvement over the baselines by 3.31 BLEU, 4.79 ROUGE-L on the original dataset, and 0.74 BLEU, 1.53 ROUGE-L on our new dataset. In addition, it is found that pre-training can effectively improve the model performance on the small dataset. Figure \ref{fig3} 
illustrates that the process of pre-training can help decrease perplexity and prevent overfitting. However, it does not work on the large dataset which shows that our model has the ability of learning the general expression of dictionary definitions even without pre-training when there is sufficient data (Model H vs. I).

\textbf{Human Evaluation} In addition to automatic evaluations, we also leverage manual evaluations to enhance the reliability of the evaluations. We randomly select 100 entries from the test set of Oxford-2019 dataset and ask 5 students to annotate. Candidate definitions include the definitions generated by our model (\textbf{Semantics-Generator}) as well as the baselines (\textbf{S+G+CH}, \textbf{S+I-Attention}) and are shuffled for each student. The human annotators evaluate these samples on two aspects: the overall quality and the readability. The quality is rated from 1 to 5 points as: 1) complete wrong, 2) correct theme with wrong information, 3) correct but missing parts, 4) some minor flaws, 5) correct. Readability is also scored from 1 to 5 points according to the syntactic structure, fluency and difficulty of the definition. The average scores reported in Table \ref{table4} show that our model performs the best among all baselines regardless of considering context or not.  

\begin{table}[h]
	\begin{tabular}{lll}
		\toprule[1.2pt]
		\textbf{Model}               & \textbf{Quality} & \textbf{Readability} \\ \hline
		\textit{S+G+CH}              & 2.03    & 3.33        \\
		\textit{S+I-Attention}       & 2.23    & 3.37        \\
		Semantics-Generator & \textbf{2.96}    & \textbf{3.77}        \\ \bottomrule[1.2pt]
	\end{tabular}
	\caption{Averaged human-annotated scores on randomly sampled 100 entries from Oxford-2019 dataset.}
	\label{table4}
\end{table}

\subsection{Model analysis}

\begin{figure}[h]
	\centering
	\includegraphics[scale=0.45]{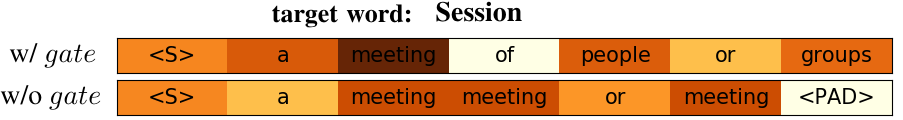}
	\caption{Token representation with gated function or not for generating the definition of ``Session''. The model with gated function pays more attention to the content words and generates more fluent definition.}\label{fig6}
\end{figure}

\textbf{Contributions of gated function and initial hidden state of decoder.} We study the effect of the gated function and the initial hidden state of the decoder on our model. First, we remove the gated function and plot the result in Figure \ref{fig4}. Removing the gated function leads to the worse metrics on both datasets, which indicates that the quality of the generated definition can be effectively improved by controlling how much information flows in at each decoding step. We also plot the average token representation of the model with gated function or not when generating the definition of the target word \emph{Session} in Figure \ref{fig6}, which shows that the gated mechanism makes the model pay more attention to the content words rather than function words. Compared without gated mechanism, the second generated word, \emph{meeting}, is given more importance, so that the model generates the correct definition and avoids the problem of repeating pattern.

Then we try to initialize the hidden state of the decoder with varied information to verify the effectiveness of the explicit signal. Figure \ref{fig5} shows that the performance of the model \textbf{Semantics-Generator} ($\boldsymbol{v}^{*}+\boldsymbol{v}_{c}$) which is provided the combination of pretrained word embedding $\boldsymbol{v}^{*}$ and context embedding $\boldsymbol{v}_{c}$ outperforms others. Pretrained word embeddings are trained on large corpora and therefore contain a wealth of semantic information that allows the decoder to obtain a stable representation during the decoding process. However, since it is context-independent, all of the meanings of a word are fused in a vector, and providing additional context information can help the decoder generate more relevant definitions.

\textbf{The effect of embeddings.}
To analyze whether additional embeddings have an impact on the model performance, we perform an ablation analysis on the target word embeddings. The results are shown in Table \ref{table5}. Removing character-level embeddings leads to 0.07 BLEU drop and 0.56 ROUGE-L drop on the full original Oxford dataset, which indicates that character-level embeddings capturing the morphological information give a positive impact to the model performance. Furthermore, removing ELMo embeddings has a strong impact on each evaluation metric, especially on the \textbf{Seen} words dataset. It shows that the contextualized embeddings provided by ELMo can effectively distinguish the meanings of the same word with different contexts, which is helpful to avoid polysemy.

\begin{table}[h]
	\begin{footnotesize}
		\begin{tabular}{p{2.1cm}lll}
			\toprule[1.2pt]
			\textbf{Model}       & \multicolumn{1}{c}{\textbf{Full}} & \multicolumn{1}{c}{\textbf{Seen}} & \multicolumn{1}{c}{\textbf{Unseen}} \\ \hline 
			W2V+CH+ELMo  & \textbf{16.00/21.94}              & \textbf{16.53/21.98}              & \textbf{13.38/21.70}                \\
			W2V+ELMo    & 15.93/21.38              & 16.50/21.70              & 13.01/20.99                \\
			W2V         & 14.16/18.83              & 14.55/18.72              & 12.21/19.37                \\ \toprule[1.2pt]
		\end{tabular}
	\end{footnotesize}
	
	\caption{The ablation study on the additional embeddings, experiments on original Oxford Dataset (BLEU / ROUGE-L).}
	\label{table5}
\end{table}

\begin{table*}[]
	\begin{tabular}{|l|p{5.4cm}|l|p{7.5cm}|}
		\hline
		\textbf{Word}                   & \textbf{Context}                                                                                                & \textbf{Model}               & \textbf{Definition}                                                                                    \\ \hline
		\multirow{4}{*}{check} & \multirow{4}{5.5cm}{\#1 We got our bill, paid the \textbf{check}, and made our way enthusiastically to Billy's bakery.} & S+G+CH              & a small quantity of something                                                                 \\ \cline{3-4} 
		&                                                                                                        & S+I-Attention       & a piece of thing that is made by a person                                                     \\ \cline{3-4} 
		&                                                                                                        & Semantics-Generator & a sum of money paid for a particular person                                                   \\ \cline{3-4} 
		&                                                                                                        & Ground Truth        & the bill in a restaurant                                                                      \\ \hline
		\multirow{4}{*}{check} & \multirow{4}{5.5cm}{\#2 The mark is only awarded to organizations that pass regular quality \textbf{checks}.}           & S+G+CH              & a small quantity of something                                                                 \\ \cline{3-4} 
		&                                                                                                        & S+I-Attention       & to be the control of a particular substance                                                   \\ \cline{3-4} 
		&                                                                                                        & Semantics-Generator & examine or inspect something                                                                  \\ \cline{3-4} 
		&                                                                                                        & Ground Truth        & an examination to test or ascertain accuracy, quality, or satisfactory condition   \\ \hline      
	\end{tabular}
	\caption{Generated definitions for word ``check'' in Oxford-2019 test set.}\label{table6}
\end{table*}
\begin{figure*}[t]
	\centering
	\quad
	\subfigure[Number of senses of the target word.]{
		\includegraphics[width=8.3cm]{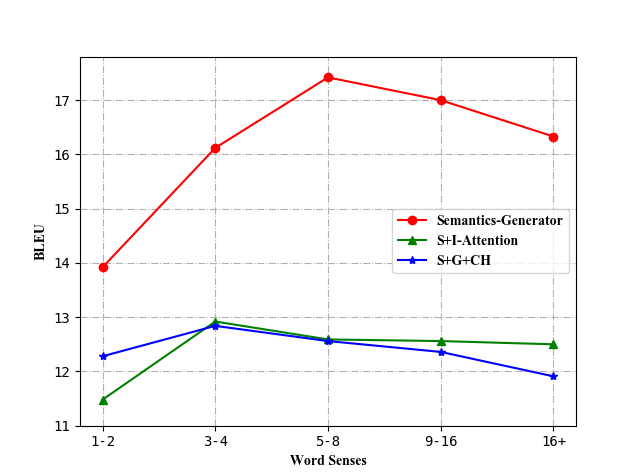}\label{fig7a}
	}
	\quad
	\subfigure[Length of the context of the target word.]{
		\includegraphics[width=8.3cm]{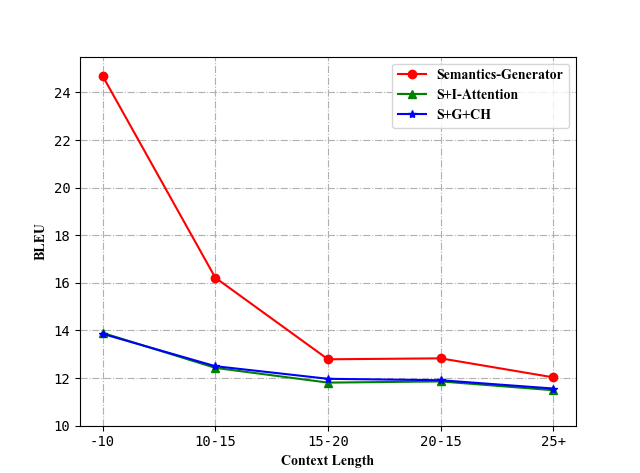}\label{fig7b}
	}
	\caption{Impact by various parameters of the target word on BLEU scores of the generated definitions. Our model Semantics-Generator achieves better performance than other models by using context information explicitly.}\label{fig7}
\end{figure*}
\subsection{Discussion}
An important ability of our model is that it can directly incorporate the context during definition generation. For qualitative analysis, we give a case in Table \ref{table6}. From the result, it can be seen that the \textbf{S+G+CH} model generates the same definition for one target word. It does not consider the varied context information, so it can only generate one of its most common senses for a word. Our model \textbf{Semantics-Generator} and \textbf{S+I-Attention} are context-dependent, while our model can utilize context information explicitly during generation by providing context embedding and ELMo embeddings for the decoder. Compared with just using the context to disambiguate the word sense, our model generates more reasonable definitions. For example, it can generate ``money'' given context \#1 with ``bill'' and ``paid''. We will further analyze in the following.

Further, we discuss the impact of some parameters of the target word on model performance \citep{ishiwatari2019learning}. We first analyze the impact of the number of senses of target words on definition generation in Oxford dataset. Figure \ref{fig7a} shows that the performance of the \textbf{S+G+CH} is significantly degraded when the target word is more ambiguous (sense > 8), which illustrates that the context plays an important role in the generation of polysemy definitions. The performance of our model is significantly better (3.83+ BLEU) than \textbf{S+I-Attention}, which indicates that providing context information directly is effective for decoding. And what happens if there is not enough context information? Such as there are few words in the context. In this situation, only the representation of the target word can help the model to understand the word sense. We analyze the impact of the context length on BLEU scores of generated definitions. Figure \ref{fig7b} shows that our model is much stronger (10.8+ BLEU) than the baseline models when the context is short (length < 10). Our model incorporates the contextualized representations (ELMo) from large corpus which can make up for the shortcomings of word embedding and provide better representation for the target word to help the model understand the corresponding sense, especially when the context is not available.

\subsection{Single Task vs. Multi-Task Models}
We evaluate the performance of our multi-task models by perplexity, BLEU and ROUGE-L scores and the result are reported in Table \ref{table7}. We train the single task model on definitions and usages respectively, and report their performance for comparison. The \textbf{Parallel-Shared} model can effectively improve the performance on both definition and usage modeling. This is reasonable because generating example sentence and definition are similar in some aspects, the multi-task setting can be seen as a form of inductive transfer to achieve a better generalization performance. For usage modeling, the \textbf{Hir-Shared-DU} performs better than the other models. Usage modeling task is more difficult which requires more complex syntax and semantic information, so supervising it at higher-level could have access to deeper representations transferred by the low-level definition modeling task. Table \ref{table8} shows some examples of generated definitions and usages. The result indicates that the semantic and syntactic information captured by the word embeddings has the ability to make example sentences for the target words.

\linespread{1.4}
\begin{table}[h]
	\begin{footnotesize}
		\begin{tabular}{p{2.75cm}p{2.3cm}p{2.6cm}}
			\toprule[1.2pt]
			\multirow{2}{*}{Model} & \multicolumn{2}{c}{Oxford-2019}                            \\ \Cline{1.2pt}{2-3}
			& \multicolumn{1}{c}{Definition} & \multicolumn{1}{c}{Usage} \\ \hline
			Semantics-Generator                    & 41.15/13.85/20.67              & 258.33/12.96/14.04        \\ \hline
			Parallel-Shared*             & 40.42/\textbf{13.96/20.81}             & 237.64/13.49/14.32        \\
			Hir-Shared-DU*                  & 42.67/13.57/20.41              & \textbf{227.71/13.51/14.47}        \\
			Hir-Shared-UD*                  & \textbf{39.67}/13.40/20.54              & 244.80/13.26/14.31        \\ \bottomrule[1.2pt]
		\end{tabular}
	\end{footnotesize}
	\caption{The performance of our models on Oxford-2019 test set(PPL/ BLEU/ ROUGE-L). Here, * means multi-task model.}
	\label{table7}
\end{table}

\linespread{1}
\begin{table*}[t]
	\begin{tabular}{|l|p{5.5cm}|p{4.3cm}|p{4cm}|}
		\hline
		\textbf{Word}      & \textbf{Context}                                                                                                                 & \textbf{Definition}                                                            & \textbf{Usage}                                   \\ \hline
		order     & Every person is hereby ordered to immediately evacuate the city of New Orleans.                                          & make a formal or authoritative request to someone                     & He was ordered to leave the room.        \\ \hline
		order     & The order of the knights templar was formed during the crusades when many knights and squires set out for the holy land. & a formal rule of law, especially in the roman catholic church       & The order of the church.                 \\ \hline
		skirt     & She was wearing a knee-length dark blue jean skirt with a front slit and a blue backless top.                            & a garment worn in the upper part of the body                          & She wore a silk skirt.                   \\ \hline
		skirt     & Along the scenic route skirting the rim we stopped at every lookout to gaze at the fantastic scenery.                    & go over a wide area                                                   & The road was not to be skirted.          \\ \hline
		agreement & No provision was made in the agreements for time off, sick pay, or holidays.                                           & the action of establishing a relationship between two or more parties & The agreement between the two countries. \\ \hline
		gaze      & I took my chair to the open corridor and sat there with my book, gazing at the sunset.                                  & look at a place steadily                                              & She gaze at the room.                    \\  \hline
		
	\end{tabular}
	\caption{Some examples of definitions and usages generated by our multi-task models from Oxford-2019 test set.}
	\label{table8}
\end{table*}

\section{Conclusion}
Our work focuses on improving the interpretability of word embeddings by semantics generation. We explore the definition modeling task to generate dictionary definitions of words using pre-trained word embeddings. A novel framework is proposed to generate context-dependent definitions and it outperforms other definition models on both the existing Oxford dataset and the newly collected Oxford-2019 dataset. The quantitative and qualitative analysis demonstrates the effectiveness of our model and indicates the importance of providing explicit context representation in definition generation.

In addition, we introduce usage modeling, and try to utilize word embeddings to generate usages of words. Two kinds of multi-task approaches are investigated to combine usage modeling and definition modeling, sharing the representations at different levels including parallel and hierarchical level. Our multi-task models can generate both definitions and example sentences of target words, and the experiment results show that it can take advantage of inductive transfer between tasks to achieve better performance on both definition modeling and usage modeling task.

Generating example sentences based on word embeddings is an interesting but challenging task. In future work, we plan to design more effective models with lexically constrained decoding and combine with extractive approaches to improve usage generation. In addition, we will try to apply our work to other natural language processing tasks such as machine translation and word sense disambiguation, as well as to intelligent education systems which can help people to learn unknown words effectively.

\section*{Acknowledgements}
This work is supported by the National Key R\&D Program of China under grant no.2018YFC1900804, Research Program of State Language Commission under grant no.\-YB135-89.

\bibliographystyle{apalike}
\bibliography{expert}

\end{document}